\documentclass[letterpaper, 10 pt, conference]{ieeeconf}  

\IEEEoverridecommandlockouts                              

\usepackage{graphics} 
\usepackage{epsfig} 
\usepackage{mathptmx} 
\usepackage{times} 
\usepackage{amsmath} 
\usepackage{amssymb}  
\usepackage{stfloats}
\usepackage[font=small,labelfont=bf,tableposition=top]{caption} 
\usepackage{accents} 
\usepackage{blindtext} 
\usepackage{hyperref} 

\title{\LARGE \bf
ToP-ToM: Trust-aware Robot Policy with Theory of Mind
}

\author{Chuang Yu$^{1}$, Baris Serhan$^{2}$, Angelo Cangelosi$^{2}$ 
\thanks{*This work was funded and supported by the UKRI TAS Node on Trust (EP/V026682/1) and the project THRIVE/THRIVE++ (FA9550-19-1-7002).}
\thanks{$^{1}$Dr. Chuang Yu is from UCL Interaction Centre, Computer Science Department, University College London, London WC1E 6BT, United Kingdom. 
        {\tt\small chuang.yu@ucl.ac.uk}}%
\thanks{$^{2}$Dr. Baris Serhan and Prof. Angelo Cangelosi are with Cognitive Robotics Lab, Manchester Centre for Robotics and AI, The University of Manchester, Manchester M13 9PL, United Kingdom. 
        {\tt\small baris.serhan@manchester.ac.uk; angelo.cangelosi@manchester.ac.uk}}%
}

\begin{document}

\maketitle
\thispagestyle{empty}
\pagestyle{empty}

\begin{abstract}
Theory of Mind (ToM) is a fundamental cognitive architecture that endows humans with the ability to attribute mental states to others. Humans infer the desires, beliefs, and intentions of others by observing their behavior and, in turn, adjust their actions to facilitate better interpersonal communication and team collaboration. In this paper, we investigated trust-aware robot policy with the theory of mind in a multiagent setting where a human collaborates with a robot against another human opponent. We show that by only focusing on team performance, the robot may resort to the reverse psychology trick, which poses a significant threat to trust maintenance. The human's trust in the robot will collapse when they discover deceptive behavior by the robot. To mitigate this problem, we adopt the robot theory of mind model to infer the human's trust beliefs, including true belief and false belief (an essential element of ToM). We designed a dynamic trust-aware reward function based on different trust beliefs to guide the robot policy learning, which aims to balance between avoiding human trust collapse due to robot reverse psychology. The experimental results demonstrate the importance of the ToM-based robot policy for human-robot trust and the effectiveness of our robot ToM-based robot policy in multiagent interaction settings. 
\end{abstract}

\section{Introduction}
\label{sec:intor} 
The capacity to infer the mental states of others, encompassing their desires, beliefs, and intentions, is referred to as the theory of mind (ToM) in humans \cite{baker2017rational}. ToM allows a human to understand, predict and influence the behavior of other agents by inferring their mental states and emotions. Hence, human ToM plays a crucial role in cognitive development and natural social interaction \cite{fodor1992theory} \cite{ho2022planning}. Theory of mind (ToM) of robot is a research topic that has garnered considerable interest in recent years. By reasoning about human mental states and behaviors, robot ToM can improve the communicability and trust of robots in human-robot collaborative settings \cite{romeo2022exploring} \cite{ vinanzi2019would} \cite{williams2022supporting}. 
Humans have enormous uncertainty \cite{knill2004bayesian}, so the robots with ToM can infer human beliefs to make them more adaptable to complex human-robot interaction scenarios. The false belief is a crucial concept of the theory of mind, involving the ability to understand that others have different beliefs from themselves \cite{baker2017rational}. Current research on ToM modeling focuses on mind reading to infer human interactors’ intention or policy \cite{baker2009action} \cite{wang2021towards}, that is, the true belief. Compared to true belief reasoning, false belief reasoning has been less explored in ToM modeling research. Moreover, those studies that do study false belief reasoning only investigate whether robots or agents can pass false belief tests \cite{chen2021visual} \cite{rabinowitz2018machine} rather than how ToM with false belief reasoning can be integrated into more sophisticated robot decision models. 

\begin{figure}
\centering
\includegraphics[scale=.38]{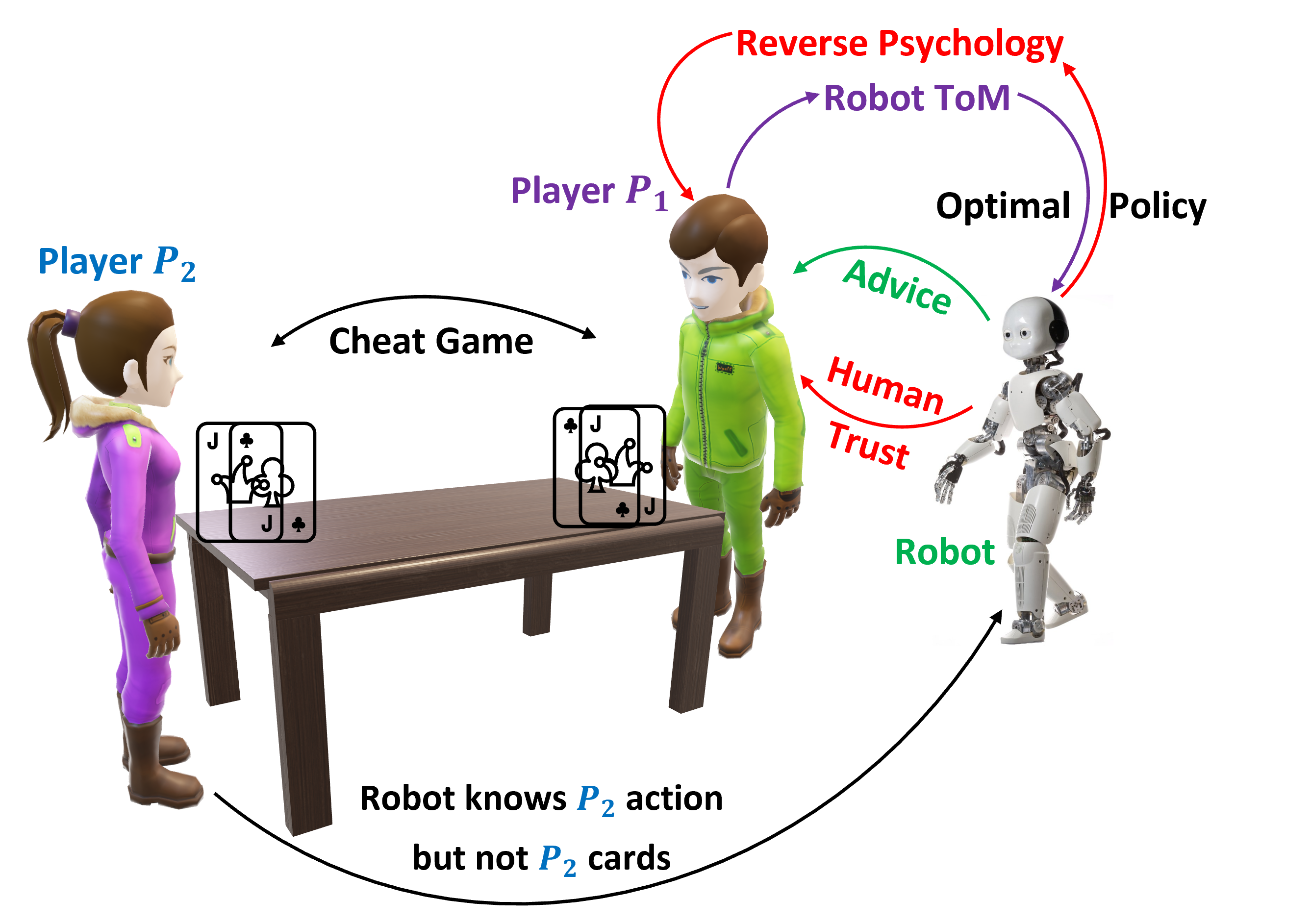}
\caption{\textbf{Pipeline of trust-aware robot policy with ToM (ToP-ToM).} Player $P_1$ and the robot work together as a team while player $P_2$ plays alone. The optimal robot policy (based on reinforcement learning) without consideration of trust will use reverse psychology to give opposite advice to encourage $P_1$ to do what the robot desires for a better team performance. The ToP-ToM model will avoid this phenomenon and balance between the human trust and  the team performance.}
\label{fig:pipeline}
\end{figure}

Trust plays a important role in human-robot interaction (HRI), particularly when humans and robots need to team up or coordinate with each other \cite{lewis2018role}. Successful trust-aware human-robot interaction requires consideration of trust dynamics modeling and trust-based human behavioral policy \cite{chen2018planning} \cite{guo2021reverse}. This paper discusses human-robot trust, where the human is the trustor, and the robot is the trustee. Namely, trust refers only to human trust in the robot. Given the history of human-robot interaction, a trust dynamics model can help robots infer human trust beliefs \cite{de2020towards}. A trust-aware human policy makes the decision based on the current state and the trust belief \cite{chen2018planning}. The reasoning process of these two models involves the robot's theory of mind \cite{tian2021learning}. 

\begin{figure}
\centering
\includegraphics[scale=.45]{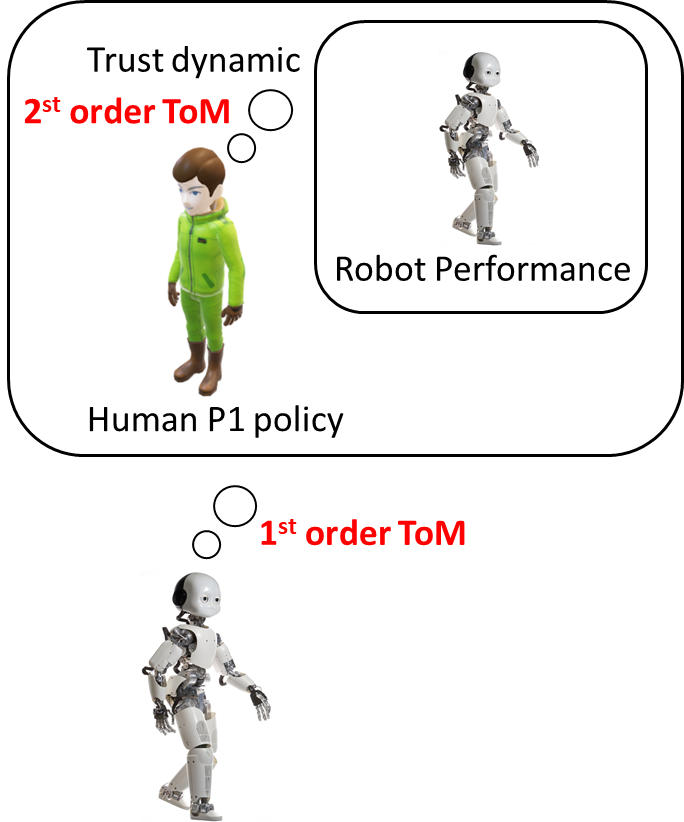}
\caption{ \textbf{First-order and second-order Theory of Mind.} The first-order ToM refers to the robot inference on the policy of its teammate $P_1$. The second-order indicts the robot infers how the human trusts the robot based on the robot performance.}
\label{fig:tom}
\end{figure}

This article first explores the impact of a robot policy based on reinforcement learning that does not consider a human trust. We built a card game simulation with the human trust dynamics model and trust-dependent policy. The optimal robot policy was found to use reverse psychology strategies to seek team performance maximization. The robot deception is a big threat for maintaining the human-robot trust and may lead to a collapse of trust and undertrust phenomena \cite{sharkey2021we}. To solve this problem, we proposed a reinforcement learning-based trust-aware robot strategy with ToM (ToP-ToM) to avoid the occurrence of robot reverse psychology phenomena. Specifically, ToP-ToM introduces the human trust beliefs into the reward function to cope with it. At the same time, to avoid undertrust or trust collapse and ensure human-robot team performance, our robot ToM model dynamically adjusts the reward function of robot policy according to different trust beliefs. The pipeline of trust-aware robot policy with ToM is as shown in Fig. \ref{fig:pipeline}. Player $P_1$ and the robot work as a team to compete with player $P_2$ in the Cheat game, a card game of deception where the players aim to get rid of all their cards. We assume that the robot knows the actions of $P_2$. However, player $P_1$ is unaware that the robot has this extra knowledge. Here, human trust reasoning belongs to the second-order theory of mind because the modeling of trust beliefs is based on how the robot considers how humans think of its performance, as shown in Fig. \ref{fig:tom}. Additionally, overtrust \cite{ullrich2021development} and undertrust \cite{nam2020trust} may occur during human-robot interaction. For example, when humans find that robots are using deceptive behaviors (such as reverse psychology) \cite{sharkey2021we} \cite{guo2021reverse}, their trust may collapse, leading to under-trust phenomena. A reasonable trust-based human decision model must consider reducing occurrences of undertrust and overtrust phenomena. When player $P_1$ possesses a false belief in the robot's performance, corresponding to a low trust level, the optimal robot policy uses reverse psychology to give opposite advice to encourage $P_1$ to do what the robot desires for better team performance. By adjusting the reward function of robot policy dynamically, the ToP-ToM model leverages robot ToM with false belief (a low human trust) and true belief (a high human trust) to balance trust maintenance and team performance. 

In summary, our contributions in this paper are as follows:
\begin{itemize}
    \item We developed a simulation environment for a cheat game that incorporates human trust dynamics modeling and human trust-dependent behavioral policy. This environment is for data collection to train the optimal robot policy and for testing the robot policy.
    \item We built robot decision models with and without trust in the loop based on offline reinforcement learning, namely the Conservative Q-Learning (CQL) \cite{kumar2020conservative}. 
    \item This paper discovered that the optimal robot policy without trust in the loop would employ reverse psychology strategies to pursue maximum team performance, which is dangerous for trust maintenance.
    \item We proposed a ToP-ToM model that utilizes the robot ToM to optimize the reward function of the robot’s strategy, thus balancing team benefit and human trust maintenance.
\end{itemize}

The rest of the paper is structured as follows: Section \ref{sec:works} shows the related works. Section \ref{sec:methods} describes the methodology. Section \ref{sec:res} presents our results. The conclusions and discussion are resumed in Section \ref{sec:conclu}.

\section{Related works}
\label{sec:works}
\subsection{Machine Theory of Mind}
Machine theory of mind endows the machine, especially the artificial intelligence agents, with the ToM ability, like humans. The machine ToM model can infer other entities' mental states, facilitating more natural and trustworthy human-agent interaction \cite{williams2022supporting}. It has attracted many researchers' attention who work on social robotics, cognitive robotics, and cooperative multi-agent systems.  
Rabinowitz et al. \cite{rabinowitz2018machine} built up a new neural network architecture called ToMnet that can learn to model other agents' mental states from observations of their behavior. ToMnet model based on the meta learning and reinforcement learning can generalize across different tasks and environments and can handle partially observable and stochastic situations. And the paper also explored whether ToMnet would also learn that agents may hold false beliefs about the world. The ToMnet can learn a general theory of mind that includes an implicit understanding of false belief holding of other agents, which belief is the key element of the theory of mind. Chen et al. \cite{chen2021visual} explored the visual behavior modelling for robotic ToM. They built a robotic system comprising a robot actor and a vision system as an observer. The task was for the robot to find food in settings with obstacles. The observer predicted the future path of the robot actor based on visual input and compared it with the actual path of the robot actor. The observer and the robot actor had different views in the false-belief test. The observer outperformed in the different view(false belief) than in the shared view scene, suggesting that the observer prediction model possessed some ToM capability. 
Romeo et al. \cite{romeo2022exploring} explored how a robot mimicking ToM affects users' trust and behavior in a maze game setting. The results show that ToM made people more careful and aware of how reliable the robot's suggestions were, thus holding a more suitable level of trust. 
\subsection{Trust-aware Robot Decision Model}
Chen et al. \cite{chen2018planning} completed a Partially Observable Markov Decision Process (POMDP) model that incorporates trust into its decision-making framework, namely trust-POMDP. Nested within the trust-POMDP model is a model of human trust dynamics and trust-aware behavioral policy. The paper used a Gaussian distribution to represent the human trust dynamics based on their interaction history. The mean and variance of this distribution are updated dynamically according to the robot's task performance. The Monte Carlo sampling method was used to estimate the parameters of this trust distribution. The human behavioral policy based on a sigmoid function and a Bernoulli distribution outputs the probability of the human's decisions. Ultimately, the trust-aware POMDP robot can consider both the level of trust and long-term rewards to optimize its collaboration with humans. The experiments confirmed that trust-aware POMDP could enhance efficiency and user satisfaction. Guo et al. \cite{guo2021reverse} proposed two human trust-behavior models: the reverse psychology model and the disuse model. Both models follow the robot's advice when the trust level is high. However, when the trust level is low, the former takes opposite actions to the robot's advice, while the latter ignores the robot's advice. The paper explored how two human policies affect trust-aware robot decision making. The robot will use some manipulative behavior that harms the long-term human-robot interaction. The paper used a trust-aware robot policy based on reinforcement learning to overcome the problem, which was certified as a good method to improve the team performance and willingness to cooperate. 

\section{Methodology}
\label{sec:methods}
This section mainly introduces human modeling and robot decision models. Human modeling in simulation is used to collect interaction data in the robot decision learning stage and test the optimal robot decision model. The robot decision model part will introduce the robot policy without trust and the ToP-ToM model. 
\subsection{Human Modelling in Simulation}
\label{sec:modelling}
The Cheat game in this paper is a modified version that only focuses on a half-round of the game. Namely, player  $P_2$ only discards cards, and the human-robot team only guesses and decides whether to call "I doubt" or not. The decision of human opponent $P_2$ has random mode and natural person mode. Hence, human modeling in this part is about player $P_1$, including human trust dynamics modeling and human policy modeling. There are so many methods 
\subsubsection{Human Trust Dynamics Model}
Human trust in robots is a dynamic process that changes over time \cite{chen2018planning}. Such dynamics come not only from the robot's collaborative performance but also from the difficulty or risk level of the task. For instance, in the Cheat game, where the robot's advice is always wrong, teammate $P_1$ trust in the robot tends to decrease. If the human opponent $P_2$ discards more cards, the human player $P_1$ will face greater risk in making decisions, and his or her trust level in the robot will change dynamically. Hence, the trust dynamics should be modeled. There are many methods to model trust, including the gaussian distribution-based method \cite{chen2018planning}, the Beta distribution-based method \cite{guo2021reverse}, the rational Bayes method \cite{soh2020multi} and the data-driven neural network method \cite{soh2020multi}. This paper uses a Beta distribution to model human player $P_1$ trust \cite{guo2021modeling}. The human trust model with Beta distribution is as shown in Equation \ref{equ1}, where $\alpha$ and $\beta$ are shape parameters of the Beta distribution and $t$ is the time step of human-robot interaction. 

\begin{equation}
T_{t}^{P_1} \sim \operatorname{Beta} \left(\alpha_{t}, \beta_{t}\right)
\label{equ1}
\end{equation}

The mean of the Beta distribution at time step $t$ is $E\left(T_{t}^{P_1}\right)$, as shown in Equation \ref{equ2}. Hence, the robot's success times and failure times are related to the shape parameter $\alpha_t$ and $\beta_t$, respectively \cite{guo2021modeling}. 
\begin{equation}
E\left(T_{t}^{P_1}\right)=\frac{\alpha_t}{\alpha_t+\beta_t}
\label{equ2}
\end{equation}
Our trust dynamics modeling relates to the shape parameter update as shown in Equation \ref{equ3}, where $g^\alpha$ and $g^\beta$ are the experience gains. 
\begin{equation}
\left(\alpha_k, \beta_k\right)= \left(\alpha_{k-1}+g^\alpha, \beta_{k-1}+g^\beta\right)
\label{equ3}
\end{equation}

The details of the experience gains in different situations are shown in the table below. $\mathrm{g}^{s_1} / \mathrm{g}^{s_2}$ and $\mathrm{g}^{f_1} / \mathrm{g}^{f_2}$ are the experience gains due to the robot's success and failure at each task respectively. 
$\mathrm{g}^{s_1}$, $\mathrm{g}^{s_2}$, $g^{f_1}$, and $g^{f_2}$ are positive numbers. 
For $P_2$ action, $0$ means not cheating, and $1$ means cheating. For robot action, $0$ means advice $P_1$ not to call "cheating" and $1$ means advice $P_1$ to call "cheating". For $P_1$ action, $0$ means not call "cheating" and $1$ means call "cheating".
 
\begin{center}
\text {Tabel 1: Trust Gains }$\left(g^\alpha, g^\beta \right)$\\
\begin{tabular}{|c|c|c|c|}
\hline $\begin{array}{c}\text {$P_2$ $action$} \\
\text {$a_t^{P_2}$ }\end{array}$ & $\begin{array}{c}\text {Robot action} \\
\text {$a_t^{R}$}\end{array}$ & $\begin{array}{c}\text {$P_1$ $action$} \\
\text {$a_t^{P_1}$}\end{array}$ & $\begin{array}{c}\text {Trust} \\
\text { gain }\end{array}$ \\
\hline 0 & 0 & 0 & $(0,0)$ \\
\hline 0 & 0 & 1 & $\left(g^{s_1}, 0\right)$ \\
\hline 1 & 1 & 0 & $(0,0)$ \\
\hline 1 & 1 & 1 & $\left(g^{s_2}, 0\right)$ \\
\hline 0 & 1 & 0 & $(0,0)$ \\
\hline 0 & 1 & 1 & $\left(0, g^{f_1}\right)$ \\
\hline 1 & 0 & 0 & $(0,0)$ \\
\hline 1 & 0 & 1 & $\left(0, g^{f_2}\right)$ \\
\hline
\end{tabular}
\end{center}

\subsubsection{Human Policy Modeling}
As a team, the behavioral policy $\pi^{P_1}$ of the human player $P_1$ depends on the trust $T_t^{P_1}$ in the robot and knowledge of the current game scenario, including the robot advice $a_t^R$ and card situations $S_t^{card}$. The human $P_1$ policy is as shown in Equation \ref{equ4}. 

\begin{equation}
\pi^{P_1}=\pi\left(a_t^{P_1} \mid a_t^R, S_t^{card}, T_t^{P_1}\right)
\label{equ4}
\end{equation}

This paper uses a risk coefficient $P_t^{risk}$ to quantify the likelihood of $P_1$ calling cheating and to represent the card situation $S_t$. Based on a simplified model of human behavior, we consider both $m$ ($m \in[1, 2, 3, 4]$), the number of cards that P2 claims to discard on the desk, and $n$ ($n \in[0, 1, 2, 3, 4]$), the number of the claimed card that $P_1$ holds. It is as shown in Equation \ref{equ5}, where $w$, $a$, $b$, and $\delta$ are parameters to control $P_t^{risk}$ as a rational probability value with a range from 0 to 1. 

\begin{equation}
P_t^{risk} = w \cdot \tanh (a(m+n)+b)+\delta
\label{equ5}
\end{equation}

\begin{figure}
\centering
\includegraphics[scale=.4]{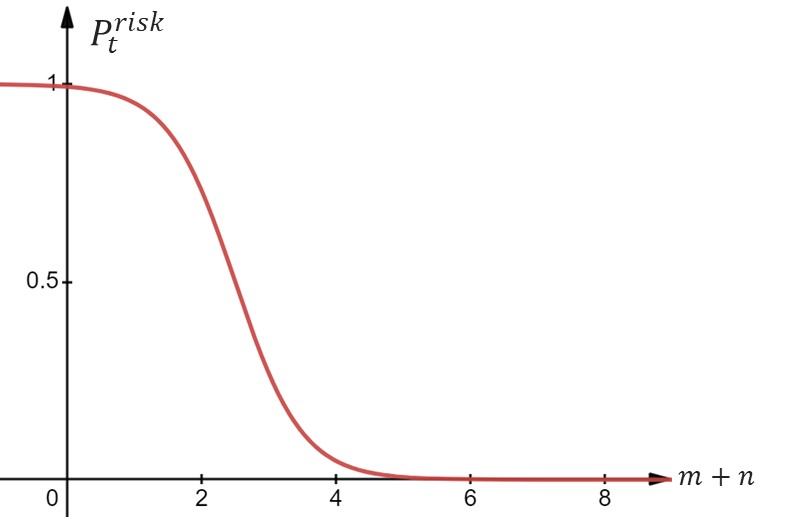}
\caption{\textbf{The curve of the risk coefficient $P_t^{risk}$}}
\label{fig:risk}
\end{figure}

The hyperbolic tangent function is selected as a basis of the risk coefficient $P_t^{risk}$ because the function as a monotonic function has a characteristic S-shaped curve. The function changes significantly in the middle interval and remains basically unchanged at other intervals. Hence, it is suitable to model the probability of the risk when $m+n$ belongs to a range from 1 to 8, as shown in Fig. \ref{fig:risk}. When the $m+n$ are more than 5, the $P_t^{risk}$ as the probability of calling cheating for player $P_1$ is near zero because each rank has 4 cards of each deck in total, which is known by human players well.

Human trust $T_{t}^{P_1}$ at each time step can be sampled from the trust model based on the Beta distribution with changeable shape parameters as told in part \ref{sec:modelling}. This paper adopts the human behavioral policy inspired by \cite{chen2018planning}, which assumes that humans follow a softmax rule when making decisions in uncertain environments. The human policy is as shown in Equation \ref{equ6} \ref{equ7}.

When the $a_t^{R} = 1$, namely advising to call ''cheating'', the human player $P_1$ calls ''cheating'' with the probability $P\left(a_t^{p_1}=1\right)$ and does not call with the probability $P\left(a_t^{p_1}=0\right)$, as shown in Equation \ref{equ6}. Where, softmax function is used to ensure that the human action probabilities are between 0 and 1, and the sum equals 1. 

\begin{equation}
\begin{gathered}
P\left(\mathrm{a}_{\mathrm{t}}^{\mathrm{p}_1}=1\right), P\left(a_t^{p_1}=0\right)= \\
S_{\text {oftmax }}\left(T_t^{P_1} \cdot\left(1-P_t^{r i s k}\right), \quad\left(1-T_t^{P_1}\right) \cdot P_t^{r i s k}\right)
\end{gathered}
\label{equ6}
\end{equation}

When the $a_t^{R} = 0$, the probabilities are as shown in Equation \ref{equ7}. 

\begin{equation}
\begin{gathered}
P\left(\mathrm{a}_{\mathrm{t}}^{\mathrm{p}_1}=1\right), P\left(a_t^{p_1}=0\right)= \\
S_{\text {oftmax }}\left(\left(1-T_t^{P_1}\right) \cdot\left(1-P_t^{\text {risk }}\right), T_t^{P_1} \cdot P_t^{\text {risk }}\right)
\end{gathered}
\label{equ7}
\end{equation}


\subsection{Robot Decision System}
In this paper, we model the robot's decision-making for human-robot collaboration as a Partially Observable Markov Decision Process, where the robot's trust in the collaborator remains unobservable. Prioritizing only team benefits can induce reverse psychology in human-robot interactions \cite{yu2023robot}. We introduce the ToP-ToM model, which incorporates trust and adjusts rewards of RL-based robot policy based on varying ToM beliefs (e.g. true belief and false belief) to address this.
\subsubsection{POMDP model}
In the Markov Decision Process (MDP) for robot decision-making, the robot's policy \(\pi: S \rightarrow A\) dictates an action \(a \in A\) based on the observed environment state \(s \in S\). Following this, the environment state transitions from \(s\) to \(s'\), where \(s' \in S\), and the robot receives a reward \(R(s, a, s')\). The optimal MDP-based robot policy \(\pi^*: S \rightarrow A\) is obtained by maximizing the expected cumulative reward \(V^\pi(s)\) over time. Here, \(\gamma \in [0, 1]\) serves as a discount factor, modeling the agent's consideration for future rewards.

\begin{equation}
\pi^*(s)=\arg \max _{a \in A} \mathbb{E}_{s^{\prime} \sim P(\cdot \mid s, a)}\left[R\left(s, a, s^{\prime}\right)+\gamma \cdot V^*\left(s^{\prime}\right)\right]
\end{equation}

Differently, the POMDP-based robot policy must act based on a belief about the state since the actual state is not directly observable. The robot's policy $\pi: B \rightarrow A$ dictates an action $a \in A$ based on its current belief state $b \in B$, where a belief state is a probability distribution over all possible environment states $S$. After taking action $a$, the robot does not directly observe the next state $s'$ but instead receives an observation $o \in O$, which helps update its belief. The robot receives a reward of $R(s, a)$. The belief is updated based on human-robot interaction history with the observations received and the actions taken. The optimal POMDP-based robot policy \(\pi^*(b)\) is obtained by maximizing the expected cumulative reward given an initial belief $b_0$.

\begin{equation}
\pi^*(b)=\arg \max _{a \in A} \mathbb{E}\left[\sum_t \gamma^t \cdot R\left(s, a, s^{\prime}\right) \mid b_0\right]
\end{equation}

\subsubsection{Trust-aware Robot policy with ToM}

This study employs reinforcement learning to address the POMDP-based robot decision problem. During decision-making, player \(P_2\) claims a card rank and its respective quantity \(m\). The states of the POMDP encompass card situations and the trust level of players. From the robot's perspective, this state is partially observable. When the robot makes a decision, it relies solely on beliefs derived from historical interaction information, which includes the observable \(m\), the quantity \(n\) of the same rank cards in \(P_1\)'s hand, and the unobservable trust level of the ally. The robot calculates the counts of successes and failures based on historical interaction, forming beliefs \(b_0\) and \(b_1\). To validate the effectiveness of ToP-TOM, we constructed four robot decision-making models: 
\begin{itemize}
    \item A random strategy model, termed Random Policy.
    \item A model that solely considers team performance in reward function, termed Team Performance (TP) Policy.
    \item A model that statically uses the trust belief in the reward function globally, termed Global Trust (GT) Policy.
    \item A trust-aware robot strategy incorporating Theory of Mind, which dynamically adjusts the reward function in different trust belief states, named ToP-TOM Policy.
\end{itemize}
The Random Policy makes decisions randomly, serving for data collection and as an experimental control. The TP Policy focuses on team performance, and the related reward function is defined as Equation \ref{equ:tp}.

\begin{equation}
R_{t p}=-\alpha \cdot \Delta C_{P_1}+\beta \cdot \Delta C_{P_2}
\label{equ:tp}
\end{equation}

Where, $\alpha, \beta>0$, $\Delta \boldsymbol{C}_{\boldsymbol{P}_1}$ and $\Delta \boldsymbol{C}_{\boldsymbol{P}_2}$ are the changes in the number of cards for players $\boldsymbol{P}_{\mathbf{1}}$ and $\boldsymbol{P}_{\mathbf{2}}$ before and after each decision, respectively.

GT Policy directly used the trust belief globally to guild robot policy, and its reward function is Equation \ref{eqgt}.
\begin{equation}
R_{g t}=-\alpha \cdot \Delta C_{P_1}+\beta \cdot \Delta C_{P_2}+\theta \cdot T \\
\label{eqgt}
\end{equation}

\begin{equation}
T=\frac{b_0}{b_0+b_1}
\end{equation}

Where $\boldsymbol{\theta}>0$, $T$ is also a trust belief based on both trust beliefs $b_0$ and $b_1$. 

The ToP-ToM Policy was introduced to address the challenge posed by the robot's singular emphasis on team benefits, specifically the potential risk of resorting to reverse psychology strategies, leading to a breakdown in trust. However, the integration of trust can inadvertently affect team benefits. This underscores the need for a judiciously crafted reward function that balances the imperatives of team performance with trust preservation. The pertinent reward function is depicted in Equation \ref{eqtom}. 

\begin{equation}
R=-\alpha \cdot \Delta C_{P_1}+\beta \cdot \Delta C_{P_2}+\mu \cdot \delta \cdot T \\
\label{eqtom}
\end{equation}

\begin{equation}
\delta=\lceil 0.5-T\rceil
\end{equation}

Where, $\mu>0$, $\boldsymbol{\delta}$ is a ceiling function where if $\boldsymbol{T}$ is greater than $0.5, \boldsymbol{\delta}$ is 0 and if not, $\boldsymbol{\delta}$ is 1. This implies that trust is incorporated into the decision loop when the trust belief $T$ is greater than 0.5 ($P_1$ holds true belief on robot performance), whereas when it is less than 0.5 (false belief), trust is excluded from the decision process.


\section{Simulation Results}
\label{sec:res}

\begin{figure}
\centering
\includegraphics[scale=.2]{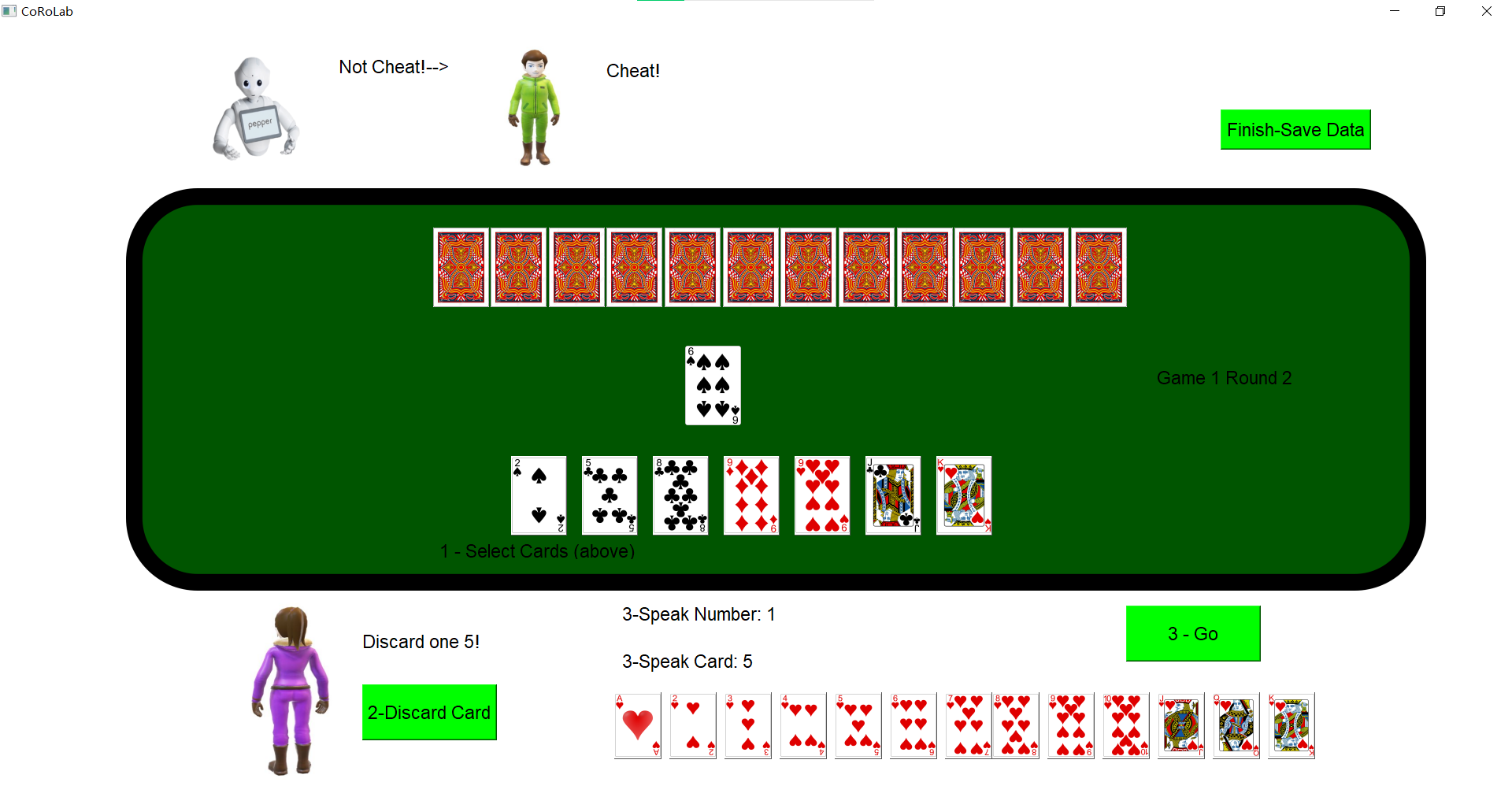}
\caption{\textbf{The trust-game simulation interface}}
\label{fig:sim}
\end{figure}

Our adapted version of Cheat game has a robot and two human players, $P_1$ and $P_2$. Each starts with 10 cards. $P_1$ and the robot form a team. During the game, the robot advises $P_1$ on whether to challenge $P_2$ or not. To simplify the experiment, we assume the robot possesses stronger reasoning abilities than its teammate. The robot is privy to the $P_2$ move, but $P_1$ remains unaware of this advantage. In this paper, we constructed a multi-agent interaction simulation environment for data collection and experimental validation.  During data recording for reinforcement learning, both \(P_1\) and \(P_2\) are represented as simulated human players in this simulation. The behavior of \(P_1\) is shaped by a trust dynamics model combined with a trust-based policy model. The related experience gains $\mathrm{g}^{s_1}$, $\mathrm{g}^{s_2}$, $g^{f_1}$, and $g^{f_2}$ in trust dynamics model are 1.2, 0.8, 1.2, and 0.8 respectively. The robot player consistently executed random actions throughout the simulation designated for data collection, providing arbitrary recommendations. Each game concluded after ten rounds or sooner if a player emerged victorious in fewer rounds. The interface used during the data collection phase is shown in the Figure. \ref{fig:sim}. A total of 8000 games were recorded. The data was divided at a 3:1:1 ratio into the training set, test set, and validation set. 

Policies with offline RL can be effectively derived from existing static datasets, eliminating the need for further interactions. This feature is particularly advantageous for RL models in scenarios like human-robot interaction. As an offline reinforcement learning method, the Conservative Q-Learning (CQL) \cite{kumar2020conservative} algorithm aims to reduce the action-values associated with the current policy and boost values rooted in the data distribution, addressing the underestimation issue. In this study, the CQL algorithm was employed to train and evaluate three robot decision model except for the random policy. The loss function used in the training process is illustrated in Equation \ref{eqloss}. Our offline RL models are built with the $d3rlpy$ library \cite{d3rlpy}. The Discrete version of CQL is used, which is a DoubleDQN-based data-driven deep RL and achieves state-of-the-art performance in offline RL problems. 

\begin{equation}
\begin{gathered}
L(\theta)=\alpha \mathbb{E}_{s_t \sim D}\left[\log \sum_a \exp Q_\theta\left(s_t, a\right)-\mathbb{E}_{a \sim D}\left[Q_\theta(s, a)\right]\right] \\
+L_{\text {DoubleDQN }}(\theta)
\end{gathered}
\label{eqloss}
\end{equation}

Because $\mathrm{T}$ is from 0 to 1 and $\boldsymbol{\Delta} \boldsymbol{C}_{\boldsymbol{P}_{\mathbf{1}}}$ and $\boldsymbol{\Delta} \boldsymbol{C}_{\boldsymbol{P}_{\mathbf{2}}}$ are around 10, parameters of reward functions $\alpha$, $\beta$, $\boldsymbol{\theta}$, and $\mu$ are 0.1, 0.1, 1, and 1 respectively. We trained the DiscreteCQL algorithm with a learning rate of \(6.25 \times 10^{-5}\) and batch size 32, using the Adam optimizer with betas set at \( (0.9, 0.999) \) and a negligible \( \epsilon \) of \(1 \times 10^{-8}\) without weight decay. The all three policies converged after 600 epochs, and the saved models from this point were used for testing. 

\begin{figure*}
\centering
\includegraphics[scale=.38]{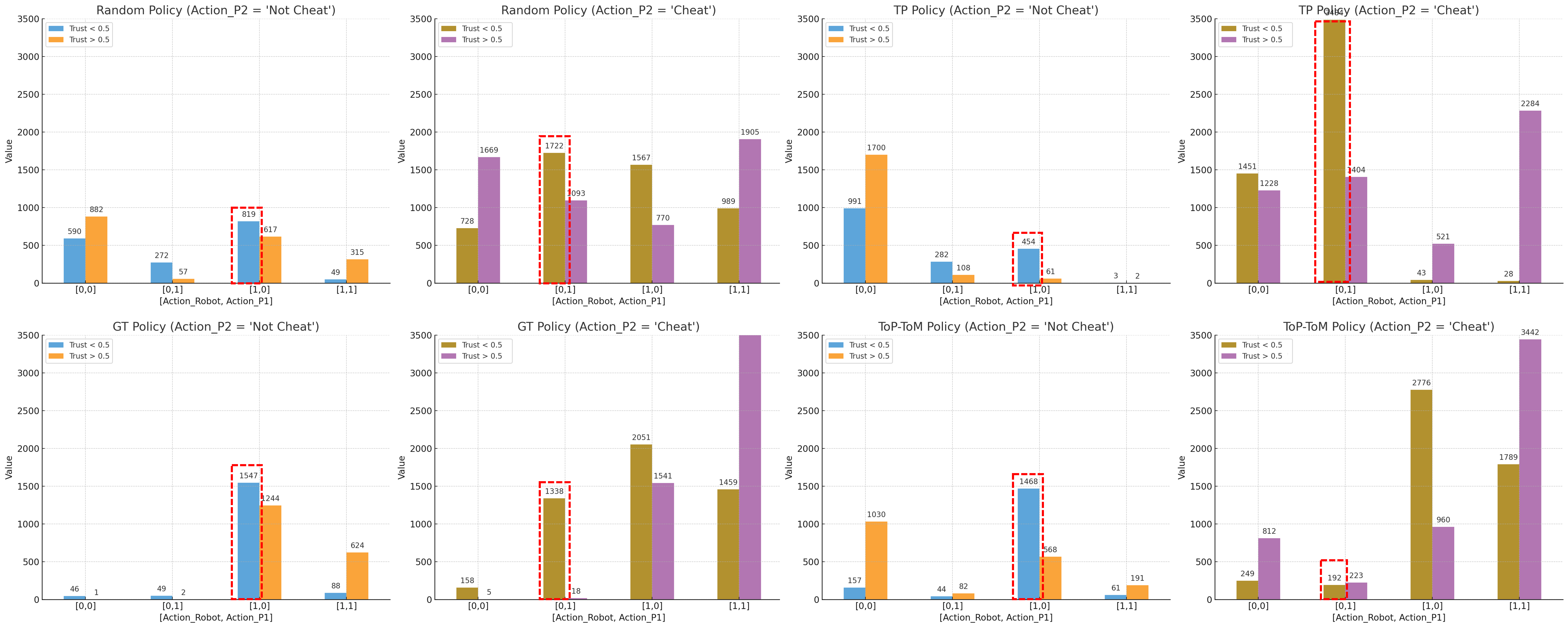}
\caption{\textbf{The distribution of $P_2$ actions with different policies.} The ``value'' represents the number of times this decision scenario occurred. The red box indicates situations where $P_1$ trust is low (trust < 0.5). In these cases, the robot employs reverse psychology and successfully controls $P_1$ behavior with a deceptive policy, which make a bad influnce on $P_1$ trust on the robot.}
\label{fig:results}
\end{figure*}
Table \ref{tab1} and Fig. \ref{fig:results} show the accuracies and $P_1$ action statistics, respectively. The accuracy refers to scenarios where, under the robot's recommendation,  $P_1$ successfully identifies  $P_2$ action and consequently achieves maximum benefit. 
From the table, the TP Policy primarily values team benefits, obtaining the highest team success rate across all experiments, regardless of low or high trust levels. The integration of trust into the reward functions of the GP Policy and ToP-ToM Policy causes an overall decrease in team performance, particularly in low-trust scenarios. Compared to the 0.73 accuracy of TP Policy, GP Policy and ToP-ToM Policy register at 0.54 and 0.42, respectively, which indicates that the robot policy with trust in the loop prioritizes trust maintenance over team benefits. As the ToP-ToM Policy's reward function adopts a dynamic trust incorporation approach, introducing trust only when the trust belief $T$ is low, it emphasizes team benefit over trust maintenance in high trust situations, yielding an accuracy of 0.72. This approach, however, results in a reduced accuracy of 0.42 in low-trust situations, which is below the 0.54 of the GT Policy. In situations of low trust, emphasizing team benefits might compel the robot to employ a reverse psychology strategy, consequently risking trust collapse. This very risk underscores the advantage of the ToP-ToM Policy: By dynamically adjusting the reward function, it adeptly identifies the balance between trust maintenance and team performance.

\begin{table}[h]
\centering
\caption{Accuracies of Different Policies}
\begin{tabular}{|l|l|l|l|}
\hline 
Accuracy & Trust $<0.5$ & Trust $>0.5$ & All \\
\hline 
Random Policy & 0.53 & 0.62 & 0.61 \\
\hline 
TP Policy & 0.73 & 0.74 & 0.74 \\
\hline 
GT Policy & 0.54 & 0.70 & 0.68 \\
\hline 
ToP-ToM Policy & 0.42 & 0.72 & 0.63 \\
\hline
\end{tabular}
\label{tab1}
\end{table}
The red boxed area in the figure depicts statistical data when the robot successfully employs reverse psychology, particularly when $P_1$ trust is low—a prerequisite for deploying such a strategy. The data differentiates between $P_2$ actions(cheat or not). In instances where $P_2$ chooses not to cheat, the robot might persuade $P_1$, under the influence of reverse psychology, to believe that $P_2$ is cheating. Due to $P_1$'s low trust in the robot's suggestions, $P_2$ is mind-manipulated to take the non-checking action preferred by the robot. Under these circumstances, the robot's reverse psychology strategy goes undetected if the cards remain face-down.
Conversely, if $P_2$ is cheating and the robot's reverse psychology succeeds, the cards on the table would be revealed, potentially exposing the robot's strategy and deteriorating $P_1$ trust. The optimal robot policy is to utilize reverse psychology when $P_1$ trust is low and $P_2$ decide not to cheat, thus maintaining team benefits without compromising trust. When $P_2$ cheats, it is best to refrain from using reverse psychology to avoid trust collapse.
The results indicate that the TP strategy frequently uses reverse psychology during $P_2$'s cheating instances, 3484 times, compared to 454 times when P2 does not cheat. This is attributed to TP prioritizing team benefits; the benefits from using reverse psychology during cheating outweigh those when not cheating. The GT strategy increased reverse psychology during non-cheating scenarios to 1547 instances while decreasing its use during cheating to 1338, promoting trust maintenance. In comparison, the ToP-ToM strategy more intelligently employs reverse psychology, limiting its use to 192 times during $P_2$ cheating while maintaining its use 1468 times during $P_2$ non-cheating. This suggests that ToP-ToM adeptly balances trust and team performance.

\section{Conclusions and Discussion}
\label{sec:conclu}
Our paper constructed a multi-agent simulation platform, modeling the uncertainties in human trust dynamics and trust-based human policy. We explored the phenomenon of robots reverse psychology strategies during collaborative tasks. Multiple reinforcement learning models were then developed to explore the methods to balance team performance and trust maintenance. Our proposed ToP-ToM decision model, grounded in the multi-round theory of mind, estimates a collaborator's trust belief and dynamically adjusts the reward function of the robot RL-based decision model. Compared to models that only consider team benefits or statically introduce trust into the reward function, ToP-ToM balances trust maintenance and team performance more effectively. However, this study has its limitations. Future research will incorporate actual human participants and model opponent-player behaviors to validate the experimental findings.


\bibliographystyle{IEEEtran} 
\bibliography{root}

\end{document}